\documentclass[10pt,twocolumn,letterpaper]{article}
\usepackage{framed,multirow}
\usepackage{cvpr}
\usepackage{times}
\usepackage{epsfig}
\usepackage{graphicx}
\usepackage{amsmath}

\usepackage{amssymb}
\usepackage{xcolor}
\usepackage{subcaption}
\usepackage[ruled,linesnumbered]{algorithm2e}

\usepackage{framed,multirow}

\usepackage[breaklinks=true,bookmarks=false]{hyperref}

\cvprfinalcopy 

\def\cvprPaperID{****} 

\begin{document}

\title{Generative Model-driven Structure Aligning Discriminative Embeddings for Transductive Zero-shot Learning}

\author{Omkar Gune$^*$, Mainak Pal$^{1**}$, Preeti Mukherjee$^{1**}$, Biplab Banerjee$^*$, and Subhasis Chaudhuri$^*$\\
\small $^*$IIT Bombay, India and $^{**}$Jadavpur University Kolkata, India \\
\tt\small guneomkar@ee.iitb.ac.in, mainak.pal08@gmail.com, preetimukherjee08@gmail.com, \\ \tt \small getbiplab@gmail.co, sc@ee.iitb.ac.in
}
\maketitle
\begin{abstract}
   Zero-shot Learning (ZSL) is a transfer learning technique which aims at transferring knowledge from \textit{seen} classes to \textit{unseen} classes. This knowledge transfer is possible because of underlying semantic space which is common to seen and unseen classes. Most existing approaches learn a projection function using labeled seen class data which maps visual data to semantic data. In this work, we propose a shallow but effective neural network-based model for learning such a projection function which aligns the visual and semantic data in the latent space while simultaneously making the latent space embeddings discriminative. As the above projection function is learned using the seen class data, the so-called projection domain shift exists. We propose a transductive approach to reduce the effect of domain shift, where we utilize unlabeled visual data from unseen classes to generate corresponding semantic features for unseen class visual samples. While these semantic features are initially generated using a conditional variational auto-encoder, they are used along with the seen class data to improve the projection function. We experiment on the both inductive and transductive setting of ZSL and generalized ZSL and show superior performance on standard benchmark datasets AWA1, AWA2, CUB, SUN, FLO, and APY. We also show the efficacy of our model in the case of extremely less labeled data regime on different datasets in the context of ZSL.
\end{abstract}

\section{Introduction}
\label{sec:intro}
\footnote{Equal contributions}
Large annotated data and advances in computing power have been two agents behind the recent success of deep learning. Many of the deep learning models have achieved performance even comparable with the human. However, collecting such data is a tedious and time-consuming task. To overcome the challenge of annotating the large data while still acquiring the impressive performance on visual recognition task, researchers have moved to other methods such as transfer learning. On the other hand, it is worth noting that humans tend to learn and recognize objects using a small amount of training data and in some cases, \textit{no} training data as well. The later case of using no training data to recognize previously \textit{unseen} object classes is made possible by associating information from different domains such as visual and textual. For example, if a child has not seen a zebra before, he/she can still recognize it when told about its description that a zebra is a horse-like animal with black and white stripes. In such scenarios, the child uses Zero-shot Learning (ZSL) to recognize previously unseen zebra using potentially already seen animals (such as dog, cat, elephant, etc.) and textual description of both seen and unseen animals.

Formally, ZSL aims at identifying previously unseen object classes using previously seen object classes and semantic side information. This transfer learning from seen to unseen is possible because of common semantic space that can represent both seen and unseen object classes. Examples of such semantic information can be textual description, attribute vector or word vector representations which are often called as semantic embeddings, semantic prototypes, class prototypes or merely prototypes. 

\begin{figure}
\centering
\includegraphics[scale=0.4]{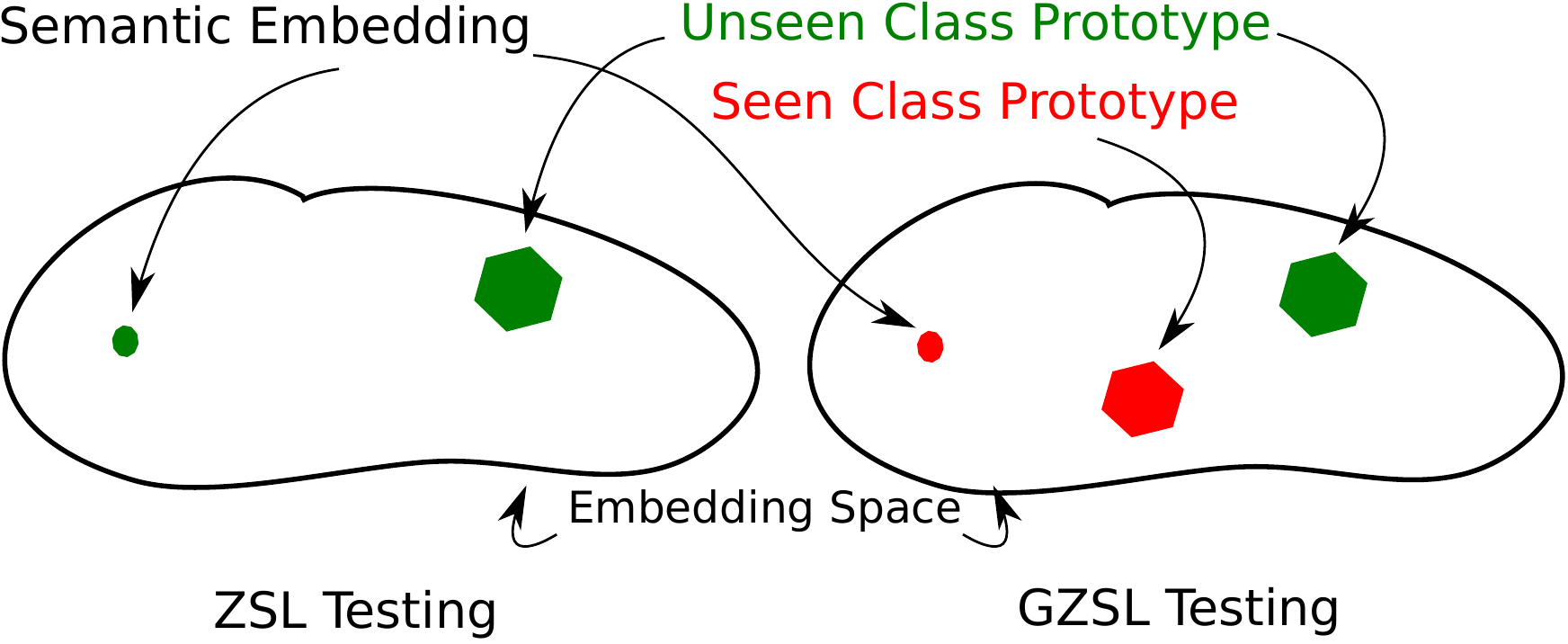}
\caption{Illustrative figure showing the domain shift in GZSL testing. If a ground truth unseen class of a test sample is closer to one of the seen classes in embedding space, the embedding of the test sample can be closer to the seen class prototype than the actual unseen class prototype resulting in miss-classification due to nearest neighbor criterion.}
\label{fig:domain-shift}
\end{figure}

A typical ZSL model learns a mapping function from visual to semantic space using labeled seen class visual-semantic data during training. During testing, the learned mapping is applied to a visual sample (e.g., image) to get its corresponding semantic embedding. This embedding is then subsequently compared with ground truth prototypes of unseen classes using nearest neighbor criterion to infer the class label. While ZSL allows only unseen class samples during testing, a practical situation would be when seen, and unseen class samples appear during testing. Such a setting is termed as Generalized Zero-shot Learning (GZSL). In case of GZSL, seen and unseen class prototypes are made available with which nearest neighbor comparison of embedding of a test sample is carried out to infer the label. Typically, the model performs poorly in case of GZSL due to inherent bias (domain shift) (\cite{tmv}) towards seen class data. This is illustrated in Fig.\ref{fig:domain-shift}. If for a given test sample from unseen class, its unseen class prototype is closer to the prototype of one of the seen classes, embeddings of the test sample can likely be closer to the prototype of that seen class than the actual unseen class prototype. Such a situation is due to model training on only seen class data which makes the model biased towards the seen class data.

As aforementioned, early ZSL techniques make use of regressor mapping from visual to the semantic domain. Such methods perform poorly due to the unbounded nature of the embedding/mapping function. Further, as pointed in (\cite{hubness1,hubness2,hubness3}) they also suffer from hubness issue. It is to be noted that visual and semantic spaces are different and may have different class neighborhood structures as well. Few ZSL techniques (\cite{tmv, lad}) use latent space as embedding space in which both visual and semantic data are projected. Latent space has improved the performance, but a little attention has been given towards making it discriminative. The discriminative property especially comes handy in case of fine-grained object classes. A triplet-loss based metric learning has been explored in (\cite{triplet}) to improve the discriminative nature of latent embeddings which requires appropriate mining of triplets.

Furthermore, a single prototype per class also poses difficulty in learning mapping function as it does not capture intra-class variance. As visual and semantic spaces are different, modeling the intra-class variance in semantic space using corresponding intra-class variance in visual space is a non-trivial task (\cite{rkt}). The aforementioned tasks are not addressed explicitly in (G)ZSL literature. Also note that occlusion, different lighting conditions, and viewing angles can result in large intraclass variance for a given class in visual space in which case, not all the attributes representing the given class would be present. We address the above issue using a shallow but effective neural network-based model. The domain shift or bias in the learned model due to training on seen class data is further reduced by using unlabeled visual data from unseen classes in \textbf{transductive} setting. We harness the power of generative model using variational auto-encoder (VAE) to generate per sample attributes for unseen classes which address the issue of sparsity in semantic space (\cite{tmv}). Training of proposed model using labeled seen data and unlabeled unseen class data with generated unseen class prototypes helps in mitigating the bias towards seen classes. Overall following are the main contributions of this work \footnote{The work in the inductive setting has been published in British Machine Vision Conference, Newcastle upon Tyne, 2018. }. 
\begin{itemize}
\item We train a shallow neural network-based model to learn a latent space where visual and semantic class data is projected. We make use of class-encoder (\cite{class-encoder}) and softmax based classifier, which makes latent embeddings to have smaller intra-class variance and larger inter-class separation, respectively. 
\item Visual and semantic space are aligned by aligning visual prototypes and semantic prototypes in the latent space. As the latent space is learned using auto-encoder like structure, latent visual and semantic embeddings perform better than original visual or semantic space features. 
\item We address the aforementioned issue of prototype sparsity for unseen classes by generating per sample semantic features for visual samples of unseen classes. We train a conditional VAE (CVAE) on labeled seen class data and use it to generate semantic features for unlabeled visual data of unseen classes.  
\item The model bias in the inductive setting is addressed by augmenting training data in an inductive setting with i) unlabeled visual data from unseen classes, and ii) generated per sample semantic features for unseen classes. In the initial step of model training, unlabeled visual samples are assigned pseudo-labels using K-means clustering, which are further pruned by classifier in each training iterations. We refer the overall transductive training as the self-taught learning (\cite{self-taught-learning}). 
\item We experiment in both inductive and transductive setting for ZSL as well as GZSL. Our results on the standard benchmark datasets AWA1 (\cite{awa}), AWA2 (\cite{good-bad-ugly}), CUB (\cite{CUB}), APY (\cite{object-by-attributes}), FLOWER (FLO) (\cite{flo}), and SUN (\cite{sun}) show superior performance compared to the state of the art methods.
\item We also show the superior performance of our model in \textit{extremely less labeled data regime} where we train with very few samples (1\%, 5\%, 10\% etc.) per seen class. 
\end{itemize}
\section{Related Work}
\textbf{Inductive ZSL:} ZSL methods use semantic side information in the form of attributes (\cite{semantic-similarity,object-by-attributes,relative-attributes, unreliable-attributes,relative-attributes, resist-to-share}, word vector (\cite{word2vec}) representations such as (\cite{semantic-similarity,cross-modal,akata-evaluation}) and image sentence descriptions (\cite{sentence-zsl,deep-zsl}). Although attributes require manual annotations, they are more effective than the other semantic features such as word vector representations. Most of the (G)ZSL methods learn a high dimension regression function which maps visual data to semantic data or vice versa as in (\cite{deep-zsl, devise, semantic-similarity, embarrassingly, sae, multi-cue}). Some methods learn latent space as embedding space (\cite{tmv, bmvc}). Deep models have been used to realize the embedding function in (\cite{deep-zsl, deep-zsl1}). As semantic space is common to both seen and unseen classes, (\cite{mixture1,mixture2,mixture3,mixture4}) represent each unseen class in terms of seen classes. To overcome the model bias towards seen classes methods in (\cite{piyush-rai-cvpr,fgn-akata, cycle-gan} make use of generative models to generate unseen class visual features which are augmented with seen class visual features. They subsequently train the classifier on seen and generated unseen data in a supervised setting. Few non-data augmentation techniques such as in (\cite{deep-caliberation}) calibrate deep network based on the confidence of seen classes and uncertainty of unseen classes. Models in (\cite{soma-biswas,bmvc,bmvc2017,structure-alignemnt-eccv18}) learn to align the neighbourhood-class structures in the visual and semantic space while meta-learning strategy (\cite{learning-to-compare}) is also studied in the context of ZSL. The issues with dataset splits are studied in detail in (\cite{good-bad-ugly}) which also proposes new dataset split and unified evaluation metric for GZSL which we follow in our work.  
\\
\textbf{Transductive ZSL:}
Among the transductive setting, QFSL (\cite{qfsl}) learns the embedding function by mapping visual data to semantic data using labeled seen data while unlabeled unseen data is forced to project onto other points specified by unseen classes. (\cite{tmv}) Uses Canonical Correlation Analysis to align multi-view embedding in latent space which is then followed by label propagation in a multi-view hypergraph.  The model in(\cite{iitropar}) uses a weak transfer constraint for knowledge transfer from seen to unseen classes. The problem of ZSL is tackled using dictionary learning  in(\cite{dictionary}) which treats the ZSL problem as unsupervised domain adaptation problem. Propagated Semantic Transfer in (\cite{transfer-learning-in-transd}) makes use of the manifold structure of unknown/novel categories and proposes graph-based knowledge transfer. Shared Model Space (\cite{guo2016transductive}) uses attributes to allow knowledge transfer between seen and unseen classes.
\begin{figure*}
\centering
\includegraphics[scale=0.5, bb={0, 0, 100, 100}, trim={0 50mm 0 0}]{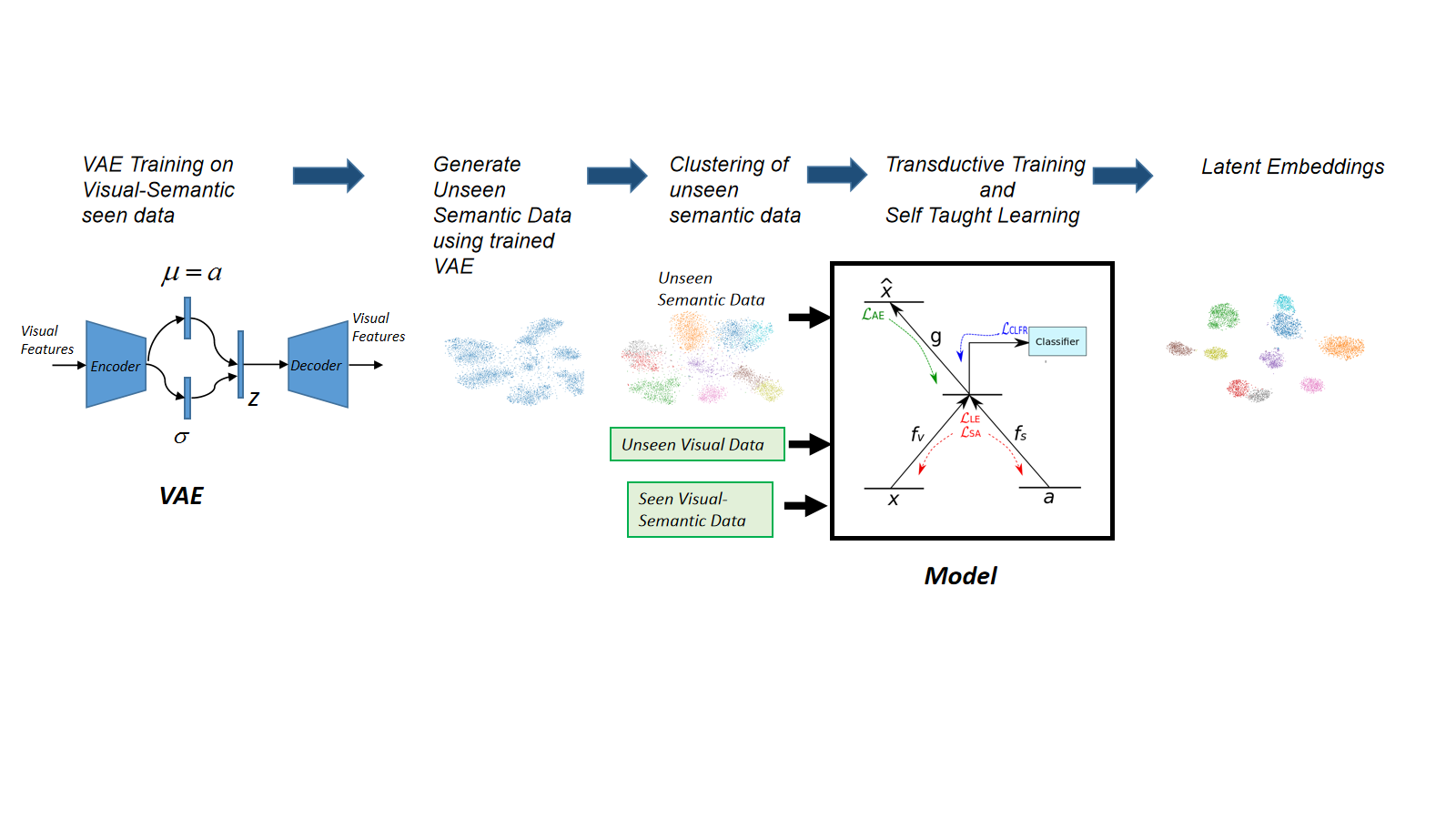}
\caption{The pipeline of the training workflow for the transductive setting. A black box outlines the NN-based model, which is used in both inductive and transductive training. The NN-based model consists of the encoder-decoder framework. During training, visual feature $\mathbf{x}$ and semantic feature $\mathbf{a}$ are projected onto common latent space. Visual feature is reconstructed back at the decoder as $\hat{\mathbf{x}}$. The classifier is learned using the latent layer features to make them discriminative. During ZSL testing, the learned visual embedding function $f_v$ and semantic embedding function $f_s$ are used to get visual and semantic latent embeddings of a test sample and unseen class prototypes. For the transductive training, a VAE is trained on seen class visual-semantic data by constraining latent distribution for a class to be Gaussian distribution with unit variance and the mean corresponding to class prototype. This trained conditional VAE is used to extract the semantic feature of unseen class visual samples. This is followed by initial clustering to assign pseudo-labels to unseen semantic features. The NN-based model is trained in the transductive setting by sampling labeled data from seen classes and pseudo-labeled data from unseen classes. Clusters on the right shows the unseen class latent embeddings during transductive training for AWA1 dataset. Best viewed in colors.}
\label{fig:model}
\end{figure*}
The proposed technique in this work differs from previous models in the following aspects. Inductive setting: First, the class-encoder (a variant of an auto-encoder)(\cite{class-encoder}) allows to have a lower intra-class variance in visual space which in turn makes latent visual embedding compact. Further, it is well known that auto-encoder provide robust latent space representations. We make use of the latent space learned by auto-encoder (class-encoder) to perform (G)ZSL. We also note that improved performance can also be attributed to the decoder part which reconstructs back visual samples from latent space. Unlike most of the existing methods which learn only forward mapping from visual to semantic or from visual and semantic to latent, we use encoder-decoder framework which has been proved to be effective due to cycle consistency (\cite{sae, cycle-gan}). Second, the classifier in the latent space which is trained on only seen classes boosts the performance by making classes in the latent space discriminative. Third, we observe that aligning projections of visual features with their corresponding class prototypes, such as in (\cite{devise,deep-zsl}), alone would not be sufficient. Hence, we also align embeddings of visual means and class prototypes in latent space. 

Transductive setting: First, we train CVAE using labeled seen class data by constraining latent codes of seen classes to come from Gaussian distribution with mean same as the corresponding class prototype. Unlike other methods in (\cite{qfsl, iitropar}) we do not use unlabeled unseen class prototypes but generate them using CVAE trained on seen data. Second, we exploit the structure in an unlabeled unseen class data using K-means clustering and assign pseudo-labels to the clusters. Third, we train the latent classifiers using data from seen and unseen classes where seen class samples are labeled using ground truth labels while unseen class samples are labeled using pseudo-labels. We iteratively train the model where pseudo-labels are updated based on the classifier outputs. We note that only the latent mean constraint is imposed on CVAE unlike (\cite{piyush-rai-cvpr}) where the superior performance of our model can be attributed to iterative pruning of labels of unseen data, latent space classifier, and a novel constraint for neighborhood alignment in the latent space. The domain shift issue is addressed by augmenting the seen data with unlabeled unseen visual data and generated unseen semantic features. The explicit consideration of domain shift or model bias and sparsity of prototypes for unseen classes demonstrates improved performance for ZSL as well as GZSL in the transductive setting. While (G)ZSL itself is considered as learning with less data, we further extend it to extremely less labeled data regime by using subsets of seen class data for training. We show that our proposed model outperforms (\cite{learning-to-compare}) in the extremely less labeled data regime. 
\section{Proposed Model}
\subsection{The problem of ZSL}
Consider a dataset $\mathcal{D}$ which contains $s$ seen classes and $u$ unseen classes.
 Let $D_S = \{\mathbf{x}_i, y_i\}_{i=1}^{n_s} \in \mathcal{X} \times \mathcal{Y}$ denote the seen data with $\mathbf{x}_i \in \mathbb{R}^d$ as the visual feature and $y_i$ as the label from $\mathcal{Y}_S = \{1,2,\cdots,s\}$ Let $D_S = D_{tr} \bigcup D_{tr}^\prime$. Here, $D_{tr}^\prime$ denotes the held out training data. Let $D_U = \{\mathbf{x}_j, y_j\}_{j=1}^{n_u}$ be the test data with $\mathbf{x}_j\in \mathbb{R}^d$ being the $j^{th}$ test sample and $y_j$ is the corresponding label from $\mathcal{Y}_U=\{s+1, s+2, \cdots, s+u\}$ such that $\mathcal{Y}_S \cap \mathcal{Y}_U = \phi$. ZSL allows the knowledge transfer between seen and unseen classes using semantic space. We denote by $\mathbf{a}_i \in \mathbb{R}^k$ the semantic class prototype for class $i$. $\mathbf{a}_i$ can be attribute vector, word vector representation or sentence description. Specifically, let $\mathcal{A}_S = \{\mathbf{a}_1, \mathbf{a}_2, \cdots, \mathbf{a}_s\}$ and $\mathcal{A}_U = \{\mathbf{a}_{s+1}, \mathbf{a}_{s+2}, \cdots, \mathbf{a}_{s+u}\}$ denote the set of seen and unseen class prototypes, respectively. Let $f_v(\cdot)$ and $f_s(\cdot)$ be the embedding functions which separately project the visual and semantic descriptors, respectively, to the common latent space. The standard ZSL problem in inductive setting aims at learning a compatibility function $\mathcal{F}(f_v(\mathbf{x}), f_s(\mathbf{a}))$ using $D_{S}$ and $\mathcal{A}_S$ such that $\mathcal{F}(\cdot)$ scores a large value when $\mathbf{x}$ and $\mathbf{a}$ belong to the same unseen class otherwise produces low values.
During testing, the label for a test sample $\mathbf{x}_t$ is estimated as follows:
\begin{equation}
\centering
\widehat{y}_t = \arg \max_{y \in \mathcal{Y}_U} \mathcal{F}(f_v(\mathbf{x}_t), f_s(\mathbf{a}_y))
\label{eq:1}
\end{equation}
In case of GZSL, a model is trained using $D_{tr}$ and $\mathcal{A}_S$ and during testing a sample can come from $D_{tr}^\prime$ or $D_U$ such that in Eq(\ref{eq:1}), $y \in \mathcal{Y}_S \bigcup \mathcal{Y}_U$.
\subsection{Inductive GZSL}
The proposed model for (G)ZSL is shown in Fig.\ref{fig:model}, which resembles the auto-encoder structure. While Fig.\ref{fig:model} illustrates the transductive workflow, we differ to it until the next section and concentrate on different losses on which model is trained in the inductive setting. We aim to learn $f_v(\cdot)$ and $f_s(\cdot)$ which project the visual and semantic data onto the common latent space, respectively, where $\mathcal{F}(\cdot)$ is evaluated to find compatibility between the visual and semantic embeddings. We build upon the encoder-decoder framework similar to auto-encoder, but the training regime is inspired by that of class-encoder (\cite{class-encoder}) for the visual branch of the model. On the other hand, we use a separate encoder network for the semantic prototypes. 

We briefly describe the standard auto-encoder (AE) model and subsequently define our loss function. An AE in its simplest form is a three-layer neural network whose goal is to reconstruct the input at the output using an encoder-decoder framework. 
Let $\mathbf{X} = [\mathbf{x}_1 \mathbf{x}_2 \cdots \mathbf{x}_{n_s}] \in \mathbb{R}^{d \times n_s}$ denote the input to AE. Here, $n_s$ is the number of samples and $d$ is the feature dimension. The encoder of AE learns mapping function $f_v(\cdot)$ by projecting $X$ onto $l$-dimension latent space ($l \ll d$) to learn latent representations $\mathbf{H} = [\mathbf{h}_1 \mathbf{h}_2 \cdots \mathbf{h}_{n_s}] \in \mathbb{R}^{l \times n_s}$. The decoder of AE  tries to reconstruct the input as $\hat{\mathbf{X}} \in \mathbb{R}^{d \times n_s}$ from their corresponding latent representations. The encoding (embedding) and decoding  functions, $f_v(\cdot)$ and $g(\cdot)$, respectively, can be realized  as follows
\begin{eqnarray}
\mathbf{h}_i  = f_v(\mathbf{x}_i) = s_f(\mathbf{W} \mathbf{x}_i)\\
\hat{\mathbf{x}}_i = g(\mathbf{h}_i) = s_g(\mathbf{W}^T \mathbf{h}_i),
\end{eqnarray}
where $s_f(\cdot)$ and $s_g(\cdot)$ represent non-linear functions such as ReLu (\cite{relu}), $\mathbf{W} \in \mathbb{R}^{l \times d}$ and bias term is neglected for representation purpose.  
The following loss is minimized while learning the AE parameters $\mathbf{W}$:
\begin{equation}
\mathcal{L}_{AE} = \sum_{i=1}^{n_s}\Vert \hat{\mathbf{x}}_i - \mathbf{x}_i \Vert_2^2.
\label{eq:loss-ae}
\end{equation}

Although AE allows learning robust latent space in an unsupervised way, class embeddings with lower intra-class variance are given by class-encoder (\cite{class-encoder}). Class-encoder exhibits a similar architecture as AE but differs in training regime. A class-encoder reconstructs the output from the different samples belonging to the same class. Let $C_x$ be the class label of two randomly picked samples $\mathbf{x}$ and $\tilde{\mathbf{x}}$. Then the training of class-encoder involves minimizing the following loss function
\begin{equation}
\mathcal{L}_{CE} = \sum_{\mathbf{x}\in X}\sum_{\tilde{\mathbf{x}}\in C_x} \Vert\hat{\mathbf{x}} - \tilde{\mathbf{x}} \Vert^2_2.
\label{eq:loss-ce}
\end{equation}
It is shown in (\cite{class-encoder}) that compact embeddings are obtained at the output of class-encoder. We believe that this also leads to compact embedding in the latent space. We observe the same experimentally (Please refer to the Experiment section).
Every class exhibits intra-class variance due to factors such as illumination changes, occlusions. In such scenarios, some of the attributes in the images may not be visible and hence should be absent from corresponding semantic representation. However, as a single prototype per class is available, it cannot reveal intra-class variability in the semantic space. Hence, to align latent embeddings obtained from visual and semantic spaces, the intra-class variance must be kept low. 
As aforementioned, we use a class-encoder to reduce the intra-class variance of the visual features in the latent space.

From a different perspective, there is no explicit constraint on the learned latent space using $f_v(\cdot)$ and $f_s(\cdot)$ that makes latent embeddings discriminative. This can become severe when dataset consists of fine-grained object classes. To overcome this issue, we learn a classifier which is trained on latent visual features to make latent space discriminative. The classifier is trained by minimizing the standard cross-entropy loss $\mathcal{L}_{CLFR}$ for $s$ seen classes. The cooperative training by the classifier results in $\mathcal{L}_{CLFR}$ to backpropagate through the encoder of the model and hence also affects the learning of embedding functions. The overall loss for the visual - latent - (visual, classifier) branch becomes

\begin{equation}
\centering
\mathcal{L}_{VIS} = \mathcal{L}_{CE} + \mathcal{L}_{CLFR}.
\end{equation}

Further, we simultaneously map class prototypes to the latent space. Given $\mathbf{a} \in \mathcal{A}_S$, $f_s(\mathbf{a})$ learns latent semantic class representations which can better be associated with the latent visual concepts than the original class prototypes. As we want latent visual and latent semantic embeddings to match, the following loss ($\mathcal{L}_{LE}$) is minimized
\begin{equation}
\mathcal{L}_{LE} = \sum_{i=1}^{n_s}\Vert \mathbf{h}_i - f_s(\mathbf{a}_{y_i})\Vert^2_2.
\label{eq:latent-loss}
\end{equation}
\textcolor{black}{Visual space is inherently different from semantic space and exhibits different class-neighborhood structures. When visual and semantic samples are projected onto latent space using non-linear embedding functions, it may be possible that class-neighborhood structures of latent visual and latent semantic samples may change. We are interested in aligning the structures in the learned latent space for visual and semantic data. This can be explained in the following way. ZSL is based on the assumption that unseen and seen classes are related. If class-neighborhood structure for seen classes is altered during training in the latent space, it may further change the relationship of seen-unseen classes in the latent space. As we have access to only seen class data, we believe that aligning structure for seen classes can help in aligning the seen-unseen class relations. We carry out this structure aligning by aligning latent representation of visual means with their corresponding latent representation of the class prototypes.} Let $\mathbf{\mu}^{i}$ denote the latent representation of the mean of visual features of class $i$. We define the structure alignment loss ($\mathcal{L}_{SA}$) as
\begin{equation}
\mathcal{L}_{SA} = \sum_{i=1}^{s} \Vert\mu^{i} - f_s(\mathbf{a}_{i})\Vert^2_2. 
\label{eqn:sa}
\end{equation}
It is to be noted that in Eq(\ref{eq:latent-loss}), we align each latent representations of the given class sample with the corresponding latent representation of the class prototype. Note that $f_s(\cdot)$ and $f_v(\cdot)$ are non-linear mappings, and hence aligning the means of the latent representations of the visual and semantic domain is different than aligning latent representations of means of visual and semantic domains. 

 The model is trained to minimize the overall loss $\mathcal{L}$ with the standard $\ell_2$ regularization $\mathcal{R}$ on the model parameters.
\begin{equation}
\mathcal{L} = \alpha_1 \mathcal{L}_{CE} + \alpha_2 \mathcal{L}_{LE} + \alpha_3 \mathcal{L}_{SA} + \alpha_4 \mathcal{L}_{CLFR} + \beta\mathcal{R}.
\label{eqn:total-loss}
\end{equation}
where $\alpha_1, \alpha_2, \alpha_3, \alpha_4$, and $\beta$ are hyper-parameters to weight the different losses.
\\
\textbf{Training and inference}: Due to non-linearity involved in $f_v(\cdot)$ and $f_s(\cdot)$, $\mathcal{L}$ is a non-convex function. The model is trained by minimizing $\mathcal{L}$ using a mini-batch gradient descent optimization strategy. For $\mathcal{L}_{VIS}$, input-output pairs are selected randomly from each of the classes in $\mathcal{Y}_S$ in each iteration of the training.
During testing of ZSL, the visual samples from $D_{U}$ and class prototypes ($\mathcal{A}_U$) of the unseen classes are projected onto the latent space using $f_v(\cdot)$ and $f_s(\cdot)$, respectively, where their projections are matched using $\mathcal{F}(\cdot)$ to assign the final label according to Eq.(\ref{eq:1}). 

In case of GZSL, visual samples (from $D_{tr}^\prime \bigcup D_U$) and semantic prototypes from seen and unseen classes (i.e., $\mathcal{A}_S \bigcup \mathcal{A}_U$) are projected onto the latent space and subsequently matched using nearest neighbor criterion to infer the final label.  
\subsection{Transductive GZSL}
Any ZSL model exhibits a bias towards seen class data as it is solely trained on it. Further, such bias poses greater difficulty in case of GZSL setting as seen class prototypes are also made available for nearest neighbor comparison with the embedding of test sample during inference stage (recall Fig.\ref{fig:domain-shift}). To illustrate this, consider that cat and tiger are present in the seen and unseen classes, respectively. During testing of GZSL cat and tiger prototypes are available for nearest neighbor comparison with the latent embeddings of the visual sample. It is likely that as cat and tiger are visually similar, the embedding of the test sample from tiger class would be closer to cat prototype than tiger prototype. This issue is also referred as \textit{domain shift} (\cite{tmv}). To alleviate this issue, we propose to use unlabeled unseen class visual samples in the transductive setting. Using additional data during training, although unlabeled, helps in finding better embedding functions. Further, one prototype per class can not capture the intra-class variance. We address the issue of prototype sparsity (\cite{tmv}) (in fact no prototypes are available for unseen classes during transductive training) for unseen classes by generating per sample semantic features for unseen class samples during training. For this, we make use of generative model. The details of this are provided below.

Generative models such as Generative Adversarial Network (GAN) (\cite{goodfellow2014generative}) and Variational Auto-encoder (VAE) (\cite{kingma2013auto}) have been successfully used in the GZSL. GAN is used in (\cite{fgn-akata}) to generate visual features of unseen classes by training conditional WGAN (\cite{improved-wgan}) on seen class data. On the other hand, conditional VAE is used in (\cite{piyush-rai-cvpr}) to generate unseen class visual samples. These methods then learn a supervised classifier which is trained on seen class data and synthesized unseen class data. While the aforementioned model synthesize the visual samples, we leverage the generating models to generate per instance semantic features.
\begin{algorithm}[t]
  \caption{Training in Transductive GZSL Setting.}
  \SetAlgoLined
  Train CVAE on seen data (visual-semantic) using constraint in Eq(\ref{eq:latent-constraint})\;
  Generate semantic features for unseen class visual samples using trained CVAE\;
  Get pseudo-labels for unseen visual data using K-means clustering\;
  \For {for the number of training iterations}{
    Sample the batch of labeled seen data (visual-semantic) \;
    Sample the batch of pseudo labeled unseen data (visual-semantic)\;
    Update the parameters of the GZSL model by optimizing the loss in Eq(\ref{eqn:total-loss})\;
    Reassign the pseudo-labels of unseen data using latent space classifier
   }
   \label{algorithm1}
\end{algorithm}
\\\textbf{Variational Auto-encoder:} 
Variational auto-encoder (VAE) is a neural network based generative model which is capable of generating data $\mathbf{x}$ using latent variables $\mathbf{z}$. A typical VAE consists of an encoder sub-network outputting the distribution  $q_{\psi}(\mathbf{z}|\mathbf{x})$ and decoder sub network outputting $p_{\theta}(\mathbf{x}|\mathbf{z})$ with $\psi$ and $\theta$ denoting the encoder and decoder network parameters. VAE aims at generating data $\mathbf{x}$ using decoder with input $\mathbf{z}$ sampled from prior distribution $p_{0}(\mathbf{z})$ by using encoder to approximate the posterior distribution of $\mathbf{z}$ given by $p_{\theta}(\mathbf{z}|\mathbf{x})$. This is achieved by training the VAE to maximize the lower bound on $p(\mathbf{x})$ by optimizing \begin{equation}
\mathcal{L}_{VAE} = \mathbb{E}_{\mathbf{z} \sim q_{\psi}(\mathbf{z}|\mathbf{x})}[\log p_{\theta}(\mathbf{x}|\mathbf{z})] - \mathcal{KL}[q_{\psi}(\mathbf{z}|\mathbf{x}) || p_{0}(\mathbf{z})],
\end{equation}
where the first term corresponds to reconstruction error of decoder and the second term represents the KL-divergence between the encoder distribution $q_{\psi}(\cdot)$ and $p_{0} $. The encoder network is trained to output the latent codes such that $q_{\psi}(\mathbf{z}|\mathbf{x})$ is defined as a standard normal distribution. Once the VAE is trained, by using an input $\mathbf{x}$, latent codes $\mathbf{z}$ can be sampled from the encoder distribution $q_{\psi}(\mathbf{z}|\mathbf{x})$.
\\
\noindent \textbf{Generating Unseen Semantic Features using Conditional VAE (CVAE):} In our case, we denote $p(\mathbf{x})$ by the distribution of visual samples of seen classes. 
We further incorporate the constraint on the latent space given by the VAE. We constraint that latent codes for a given input sample $\mathbf{x}_i$ are drawn from the Gaussian distribution whose mean is given by corresponding class prototype. Precisely, for $(\mathbf{x}_i, y_i)$ pair of seen class data we have 
\begin{equation}
    q_{\psi}(z_i|x_i) = \mathcal{N}(\mathbf{a}_i, \mathbf{I}),
    \label{eq:latent-constraint}
\end{equation}
where $\mathbf{a}_i$ corresponds to class $i$ prototype. It has been experimentally shown in (\cite{piyush-rai-cvpr}) that VAE trained on the seen class data can generate unseen class samples when its latent code $\mathbf{z}$ is supplied with unseen class prototypes. We take inspiration from this fact and instead generate latent codes using unseen visual data. We note that in our model latent codes are class prototypes unlike in (\cite{piyush-rai-cvpr}) where they use unstructured latent code (sampled from standard Normal distribution) along with class prototypes. Furthermore, we differ from (\cite{piyush-rai-aaai}) in the sense that, (\cite{piyush-rai-aaai}) models mean and variance of latent space Gaussian using additional functions, uses margin regularizer and additional probability sharpening criterion for the transductive setting. On the other hand, our CVAE model constrained only on latent means still outperform many of the existing models using self-taught learning during training, as explained in the next section. 

After training the CVAE, we use unlabeled visual samples from the unseen classes as the input to CVAE and extract latent code for each sample. As latent space is distributed with mean corresponding to class prototypes, we expect that latent codes obtained for unseen class samples are representative of corresponding class semantic representations. As semantic representations are generated on per sample basis, they capture the intra-class variance and overcomes the issue of prototype sparsity. We call these semantic representations as synthesized semantic features of unseen classes which are used during training of the proposed model.
\\
\noindent \textbf{Training in Transductive GZSL via Self Taught Learning:}
In transductive setting, we leverage the unlabeled unseen class visual samples and their synthesized semantic features along with the labeled seen class data. We note that for seen class data the reconstruction loss is given by Eq(\ref{eq:loss-ce}) while for unseen class data we use the reconstruction loss as given by Eq(\ref{eq:loss-ae}). Although unseen class data is unlabeled, it is still possible to exploit the structure in the unseen data (\cite{self-taught-learning}). We use K-means clustering for the unseen class visual samples. The K-means clustering allows us to give pseudo labels to visual samples of unseen classes. Note that the number of clusters is equal to the number of unseen classes. We use pseudo-labels of visual samples of unseen classes to label corresponding synthesized semantic features of the unseen classes. With unseen class samples are (pseudo) labeled we make the classifier in the latent space to have $s+u$ output nodes. During training in each iteration, we sample a batch of (labeled) seen class data along with (pseudo-labeled) unseen class data. As the unseen class labels are not ground truth labels, we update them based on the classifier prediction. Thus, in every iteration pseudo labels of unseen class samples are pruned using the labels predicted by the classifier. The unseen class data labeled using K-means clustering provides a good initialization. This self-taught learning by classifier helps in learning discriminative latent representations. The clustering using pseudo-labels given by the classifier performs better than clustering in the visual space. This is because compact latent space clustering can provide better neighborhood structure than in the visual space as structure aligned constraint is imposed on latent representations. We further explain below the superior performance of our CVAE coupled with self-taught learning by drawing an analogy with the pioneering work in self-taught learning (\cite{self-taught-learning}).
\begin{table}[t]
\caption{Descriptions of different datasets of (G)ZSL in terms of number of seen classes $|\mathcal{Y}_S|$, number of unseen classes $|\mathcal{Y}_U|$, dataset size i.e. train and  test samples ($\mathcal{D}$) and attribute dimension ($k$). }
\centering
\begin{tabular}{|c|c|c|c|c|c|c|}
\hline
\textbf{Item} & \textbf{AWA1} & \textbf{AWA2} & \textbf{CUB} & \textbf{SUN} & \textbf{APY} & \textbf{FLO} \\
\hline
$|\mathcal{Y}_S|$ & 40 & 40 & 150 & 645 & 20 & 82 \\
$|\mathcal{Y}_U|$ & 10 & 10 & 50 & 72 & 12 & 20\\
$|\mathcal{D}|$ & 30K & 37K & 11K & 14K & 12K & 8K\\
 $k$ & 85 & 85 & 312 & 102 & 64 & 1024\\
\hline
\end{tabular}
\label{table-dataset}
\end{table}

\begin{table*}
\caption{ZSL classification accuracy (\%) comparison on different datasets. Ours-AE: Proposed model with standard AE, Ours-CE: proposed model with class-encoder. $F_V$: VGG features, $F_G$: GoogleNet features. 
Sentence description is used in \textdagger. Kernelized prototypes are used in $*$. For CUB we use 500D CBoW word2vec from \cite{rkt} while $\ddagger$ uses 400D word2vec.}
\centering
\small
\setlength{\tabcolsep}{3.5pt}
\begin{tabular}{|l|c|c|c|c|c|c|}
\hline
 & \multicolumn{1}{|c|}{Visual} & \multicolumn{2}{|c|}{\textbf{AWA}} & \multicolumn{2}{|c|}{\textbf{CUB}} & \multicolumn{1}{|c|}{\textbf{SUN}}\\
\cline{3-7}
Method & Feature & \multicolumn{1}{|l|}{Attribute} & \multicolumn{1}{|l|}{word2vec} & \multicolumn{1}{|l|}{Attribute} & \multicolumn{1}{|l|}{word2vec} & \multicolumn{1}{|l|}{Attribute} \\
\hline
ConSE \cite{convex-combination}& $F_G$ & 59.0 & 53.2 & 33.6 & 28.8 & 49.6 \\
SSE \cite{semantic-similarity}& $F_V$ & 76.3 & - & 30.4 & - & -\\
ESZSL \cite{embarrassingly}& $F_G$ & 76.3 & - & 47.2 & - & 59.2\\
SPLE \cite{bmvc2017}& $F_G$ & 78.4 & 66.5 & 56.7 & 35.2$^*$ & 69.3 \\
SYNC \cite{mixture1}& $F_G$ & 72.9 & - & 54.5 & - & 62.8 \\
RKT \cite{rkt}& $F_G$ & 71.6 & 59.1 & 33.5 & 23.2 & - \\
ALE \cite{latent-embedding}& $F_G$ & 71.9 & 61.1 & 45.5 & $31.8^{\ddagger}$ & 63.7 \\
LAD \cite{lad} & $F_V$ & 82.5 & - & 56.6 & - & - \\
SAE \cite{sae}& $F_G$ & 84.7 & - & 61.4 & - & 65.2 \\
DZSL \cite{deep-zsl}& $F_G$ & \textbf{86.7} & 78.8 & 58.3 & 53.5\textdagger  & -\\
\hline
Ours-AE& $F_G$ & 83.7  &  79.8 & 61.2 &  28.7 & \textbf{69.6} \\
Ours-CE& $F_G$ & 85.0  &  \textbf{80.7} & \textbf{62.2} &  \textbf{31.0} & 68.1 \\
\hline
\end{tabular}
\label{table:zsl}
\end{table*}

\begin{table*}
\caption{Inductive ZSL and GZSL. Performance comparison using $ZSL_{Acc}$ and $H$ for ZSL and GZSL, respectively, on different datasets. Ours-CE: proposed model with class-encoder. $ZSL_{Acc}$: Average top 1 accuracy per class. S: Accuracy on seen classes (\%), U: Accuracy on unseen classes (\%), H: Harmonic mean (Settings followed from (\cite{good-bad-ugly}).)}
\centering
\small
\setlength{\tabcolsep}{3.5pt}
\begin{tabular}{|l|c|c|c|c|c|c|c|c|c|c|c|c|}
\hline
 & \multicolumn{4}{|c|}{\textbf{AWA1}} & \multicolumn{4}{|c|}{\textbf{AWA2}} & \multicolumn{4}{|c|}{\textbf{APY}}  \\
\cline{2-13}
\textbf{Method} & \multicolumn{1}{|c|}{$ZSL_{Acc}$} & \multicolumn{1}{|c|}{S} & \multicolumn{1}{|c|}{U} & \multicolumn{1}{|c|}{H} & \multicolumn{1}{|c|}{$ZSL_{Acc}$} & \multicolumn{1}{|c|}{S} & \multicolumn{1}{|c|}{U} & \multicolumn{1}{|c|}{H} &\multicolumn{1}{|c|}{$ZSL_{Acc}$} & \multicolumn{1}{|c|}{S} & \multicolumn{1}{|c|}{U} & \multicolumn{1}{|c|}{H} \\
\hline
\hline
DeViSE (\cite{devise})& 54.2 & 68.7 & 13.4 & 22.4 & 59.7& 74.7 & 17.1 & 27.8 & \textbf{39.8} & 76.9 & 4.9 & 9.2 \\
SYNC(\cite{mixture1}) & 54.0 & 87.3 & 8.9 & 16.2 & 46.6 & 90.5 & 10.0 & 18.0 & 23.9 & 66.3 & 7.4 & 13.3 \\
SJE (\cite{akata-evaluation}) & - & 74.6 & 11.3 & 19.6 & - & 73.9 & 8.0 & 14.4 & - & 55.7 & 3.7 & 6.9 \\
ALE (\cite{latent-embedding}) & 59.9 & 76.1 & 16.8 & 27.5 & 62.5 & 81.8 & 14.0 & 23.9 & 39.7 & 73.7 & 4.6 & 8.7 \\
SAE (\cite{sae}) & 53.0 & 77.1 & 1.8 & 3.5 & 54.1 & 82.2 & 1.1 & 2.2 & 8.3 & 80.9 & 0.4 & 0.9 \\
DZSL (\cite{deep-zsl}) & \textbf{68.4} & 84.7 & 32.8 & 47.3& 67.1  & 86.4 & 30.5 & \textbf{45.1} & 35.0 & 75.1 & 11.1 & 19.4  \\
PSR (\cite{soma-biswas}) & - & - & - & - & 63.8 & 73.8 & 20.7 & 32.3 & 38.4 & 51.4 & 13.5 & 21.4 \\
\hline
Ours-CE & 67.7 & 85.5 & 34.7 & \textbf{49.4} & \textbf{67.2} &  87.1 & 30.1 & 44.7 & 37.1 & 80.3 & 20.1 & \textbf{32.2} \\
\hline
\end{tabular}
\label{table:inductive-zsl-gzsl1}
\end{table*}

\begin{table*}
\caption{Inductive ZSL and GZSL. Performance comparison using $ZSL_{Acc}$ and $H$ for ZSL and GZSL, respectively, on different datasets. Ours-CE: proposed model with class-encoder. $ZSL_{Acc}$: Average top 1 accuracy per class. S: Accuracy on seen classes (\%), U: Accuracy on unseen classes(\%), H: Harmonic mean (Settings followed from (\cite{good-bad-ugly}).)}
\centering
\small
\setlength{\tabcolsep}{3.5pt}
\begin{tabular}{|l|c|c|c|c|c|c|c|c|c|c|c|c|}
\hline
 & \multicolumn{4}{|c|}{\textbf{CUB}} & \multicolumn{4}{|c|}{\textbf{SUN}} & 
 \multicolumn{4}{|c|}{\textbf{FLO}}\\
\cline{2-13}
\textbf{Method} & \multicolumn{1}{|c|}{$ZSL_{Acc}$} & \multicolumn{1}{|c|}{S} & \multicolumn{1}{|c|}{U} & \multicolumn{1}{|c|}{H} & \multicolumn{1}{|c|}{$ZSL_{Acc}$} &\multicolumn{1}{|c|}{S} & \multicolumn{1}{|c|}{U} & \multicolumn{1}{|c|}{H} & \multicolumn{1}{|c|}{$ZSL_{Acc}$} & \multicolumn{1}{|c|}{S} & \multicolumn{1}{|c|}{U} & \multicolumn{1}{|c|}{H}\\
\hline
\hline
DeViSE(\cite{devise}) & 52.0 & 53.0 & 23.8 & 32.8 & 56.6 & 27.4 & 16.9 &  20.9 & 45.9 & 68.7 & 13.4 & 22.4\\
SYNC(\cite{mixture1}) & 55.6 & 70.9 & 11.5 & 19.8& 56.3 & 43.3 & 7.9 & 13.4 & - &- & - &-\\
SJE(\cite{akata-evaluation}) &- & 59.2 & 23.5 & 33.6 & - & 30.5 & 14.7 & 19.8 & - & 74.6 & 11.3 & 19.6\\
ALE(\cite{latent-embedding}) &54.9 & 62.8 & 23.7 & 34.4 & 58.1 & 33.1 & 21.8 & 26.3 & 48.5 & 76.1 & 16.8 & 27.5 \\
SAE(\cite{sae}) & 33.3 & 54.0 & 7.8 & 13.6 & 40.3 & 18.0 & 8.8 & 11.8 & - & - & - & -\\
DZSL(\cite{deep-zsl}) & 51.7 & 57.9 & 19.6 & 29.2 & \textbf{61.9} & 34.3 & 20.5 & 25.6 & - & - & - & -\\
PSR(\cite{soma-biswas}) & 56.0 & 54.3 & 24.6 & 33.9 & 61.4 & 37.2 & 20.8 & 26.7& - & - & - & -\\
\hline
Ours-CE & \textbf{57.6} & 60.7 & 30.2 & \textbf{40.3} & 58.5 & 41.1 & 21.2 & \textbf{27.9} & \textbf{55.5} & 91.2 & 30.3 & \textbf{45.5} \\
\hline
\end{tabular}
\label{table:inductive-zsl-gzsl2}
\end{table*}
\begin{itemize}
    \item It has been shown in (\cite{self-taught-learning}) that unlabeled data can be used along with labeled data in self-taught learning way to improve the classification accuracy. In such case, (\cite{self-taught-learning}) exploits the structure in the unlabeled data in the sense that unlabeled data is represented using fewer basis vectors in the sparse coding framework. These basis vectors or attributes allow a higher-level representation of given images/visual features. A labeled data is represented using these bases and experimentally shown to perform better than the raw labeled data in the classification task.
    \item We term our transductive training framework as self-taught learning as i)we use CVAE latent space to obtain higher-level (attributes) representation of seen and unseen classes (similar to  sparse coding framework in (\cite{self-taught-learning})) ii) Similar to (\cite{self-taught-learning}) our classifier is trained on latent space embeddings (seen and unseen classes) which are modified from their corresponding visual and semantic features. (\cite{self-taught-learning}) uses a two-step process of representation learning and classification. On the other hand, our iterative pruning of pseudo-labels of unseen samples during training helps in improving the recognition performance on seen as well as unseen classes. 
\end{itemize}
The overall training of the transductive setting is given in Algorithm 1.
\section{Experiments}
\subsection{Datasets}
We consider the following standard datasets for the evaluation of the proposed inductive as well as transductive GZSL model: Animals with Attributes (AWA) (\cite{awa}), AWA2 (\cite{good-bad-ugly}),  Caltech Birds 200-2011 (CUB) (\cite{CUB}), a Pascal Yahoo (APY) (\cite{object-by-attributes}), FLO (\cite{flo}) and, SUN Attributes (SUN) (\cite{sun}). The details of these datasets are given in Table \ref{table-dataset} where AWA1, AWA2, and APY are coarse grained datasets while CUB, FLO, and SUN are fine-grained datasets. We use $d=2048$-dimension 101-ResNet (\cite{deep-zsl}) features as visual embeddings for all the datasets. We use manually annotated attributes for AWA1, AWA2, CUB, APY, and SUN. For FLO we use fine grained visual description given by (\cite{good-bad-ugly}) using (\cite{sentence-zsl}). We use the evaluation setting as proposed in (\cite{good-bad-ugly}) which is described in the next section.
\subsection{Evaluation Metric}
For ZSL, we average the fraction of correct predictions (top-1) per class denoted as $A_i$ to find average per class accuracy as given below
\begin{equation}
ZSL_{Acc} = \frac{1}{u}\sum_{i \in \mathcal{Y}_U} A_i.
\end{equation}
For GZSL, we calculate average per class accuracy for seen and unseen classes, denoted by $Acc_{S}$ and $Acc_U$, respectively, and find harmonic mean $H$ as proposed in (\cite{good-bad-ugly})
\begin{equation}
H = \frac{2 \times Acc_S \times Acc_U}{Acc_S + Acc_U}.
\end{equation}
\begin{table*}[t]
\caption{Transductive GZSL. Performance comparison of GZSL on different datasets. Ours-CE: proposed model with class-encoder. S: Accuracy on seen classes (\%), U: Accuracy on unseen classes (\%), H: Harmonic mean (Settings followed from (\cite{good-bad-ugly}).). The results in $*$ are not directly comparable as unlike Ours-CE these methods also use unlabeled unseen class prototype during training. The results of other methods are as reported by  (\cite{iitropar}).}
\centering
\small
\setlength{\tabcolsep}{3.5pt}
\begin{tabular}{|l|c|c|c|c|c|c|c|c|c|c|c|c|c|c|c|}
\hline
 & \multicolumn{3}{|c|}{\textbf{AWA1}} & \multicolumn{3}{|c|}{\textbf{AWA2}} & \multicolumn{3}{|c|}{\textbf{CUB}} & \multicolumn{3}{|c|}{\textbf{SUN}} & \multicolumn{3}{|c|}{\textbf{APY}} \\
\cline{2-16}
\textbf{Method} & \multicolumn{1}{|c|}{S} & \multicolumn{1}{|c|}{U} & \multicolumn{1}{|c|}{H} & \multicolumn{1}{|c|}{S} & \multicolumn{1}{|c|}{U} & \multicolumn{1}{|c|}{H} & \multicolumn{1}{|c|}{S} & \multicolumn{1}{|c|}{U} & \multicolumn{1}{|c|}{H} & 
\multicolumn{1}{|c|}{S} & \multicolumn{1}{|c|}{U} & \multicolumn{1}{|c|}{H} & 
\multicolumn{1}{|c|}{S} & \multicolumn{1}{|c|}{U} & \multicolumn{1}{|c|}{H} 
\\
\hline
\hline
ALE(\cite{latent-embedding}) & - & - & - & - & - & 21.7 & - & - & 30.4 &  - & - & 21.1 & - & - & -\\
GFZSL(\cite{piyush-rai-expo}) & - & - & -& - & - & 40.0 & - & - & 33.5 & - & - & - & - & - & - \\
DSRL (\cite{dsrl}) & - & - & - & - & - & 32.2 & - & - & 28.9 & - & - & 20.5 & - & - & \\
$^*$ QFSL(\cite{qfsl}) & - & - & - & 66.2 & 93.1 & 77.4 & 71.5 & 74.9 & 73.2 & 51.3 & 31.2 & 38.8 & - & - & -\\
$^*$ SABR(\cite{iitropar}) & - & - & - & 79.7 & 91.0 & 85.0 & 67.2 & 73.7 & 70.3 & 58.8 & 41.5 & 48.6 \ & - & - & - \\
\hline
Ours-CE & 87.7 & 71.2 & \textbf{78.6} & 86.1 & 71.3 & \textbf{78.0} & 65.7 & 46.4 & \textbf{54.4} & 37.7 & 25.8 & \textbf{30.6} & 76.1 & 39.3 & \textbf{51.8} \\
\hline
\end{tabular}
\label{table:transductive-gzsl}
\end{table*}
\subsection{Model Architecture}
The proposed model is shown in Fig.\ref{fig:model}. Although the proposed model is a shallow network, it achieves the state of the art performance in most of the experimental settings due to classifier in the latent space, structure aligning constraint, and encoder-decoder framework. We use fully connected neural network to implement the mapping functions described in section 3. The input and output dimension of the visual branch is $d$ with latent layer having dimension as 1000. We use $k$-dimension (dimension of prototype) fully connected layer as a input layer in the semantic branch and one intermediate layer having  750-dimensions. This is followed by the fully connected latent layer of 1000-dimensions. We use ReLu (\cite{relu} non-linearity at the latent and intermediate layers. In the inductive setting, the classifier has $s$ nodes at the output to which softmax function is applied. The latent visual representations are input to the classifier. In case of transductive setting, there are $s + u$ nodes at the output of classifier. We use the same model architecture for all the dataset but separately tune the hyper-parameters corresponding to weights of different loss terms ($\alpha_1 - \alpha_4 , \beta$). We use stochastic gradient descend based Adam (\cite{adam}) optimizer to train the model with learning rate of 0.0001 and a batch of $64$. We implement the proposed framework in PyTorch (\cite{pytorch}). We denote the performance of our model by Ours-CE where CE stands for class-encoder used in our model. 
\begin{table}
\caption{Transductive ZSL. Average top 1 per class accuracy (top-1)(\%) comparison on different datasets on the new split (\cite{good-bad-ugly}). Ours-CE: proposed model with class-encoder. The results in $*$ are not directly comparable as unlike Ours-CE these methods also use unlabeled unseen class prototype during training. The results of other methods are as reported by  (\cite{iitropar}).}
\centering
\small
\setlength{\tabcolsep}{3.5pt}
\begin{tabular}{|l|c|c|c|c|c|}
\hline
  \textbf{Method} & \textbf{AWA1} & \textbf{AWA2} & \textbf{CUB} &  \textbf{SUN} & \textbf{APY} \\
\hline
\hline
ALE (\cite{latent-embedding}) & - & 70.6 & 54.4 & 55.5 & - \\  
GFZSL(\cite{piyush-rai-expo}) &  - & 78.3 & 50.6 & \textbf{63.9} & - \\
DSRL (\cite{dsrl}) & - & 72.5 & 48.9 & 56.1 & - \\
$^*$QFSL(\cite{qfsl}) & - & 79.7 & 72.1 & 58.3  & -\\
$^*$SABR(\cite{iitropar}) &- & 88.9 & 74.0 & 67.5 & -\\
\hline
Ours-CE & \textbf{81.3} & \textbf{80.7} & \textbf{62.4} & 60.4 &  \textbf{49.9} \\
\hline
\end{tabular}
\label{table:transductive-zsl}
\end{table}
\subsection{Performance on ZSL and GZSL}
\begin{figure*} 
\centering
   \includegraphics[scale=0.5, bb={0, 0, 100, 100}, trim={20, 400, 0, 400}]{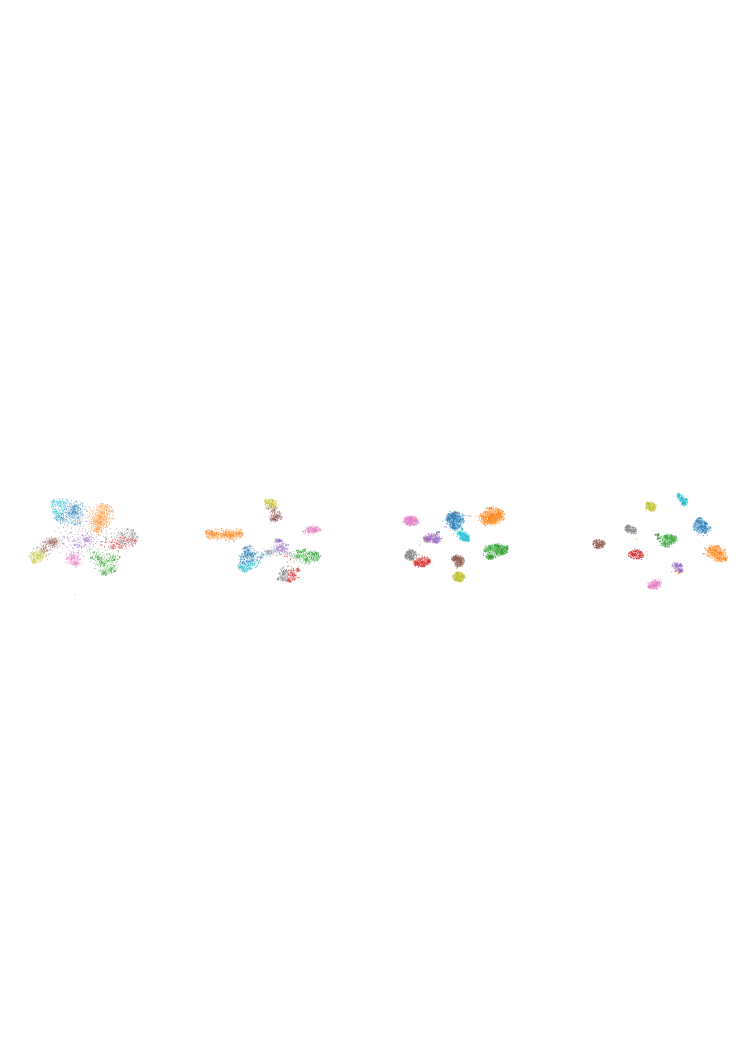}
\caption{AWA1 dataset. t-SNE plots of embeddings of unseen class visual samples in the latent space. As training progresses, self-taught learning allows forming better clusters as can be observed (from left to right). Best viewed in color.}
\label{fig:self-taught}
\end{figure*}

\noindent \textbf{Inductive Setting:}\footnote{The results (Ours-CE) in Table \ref{table:inductive-zsl-gzsl1} and \ref{table:inductive-zsl-gzsl2} on AWA1, AWA2, CUB, and SUN datasets are borrowed from our previous work in (\cite{bmvc}).}
It is clear from the Table \ref{table:inductive-zsl-gzsl1} and \ref{table:inductive-zsl-gzsl2} that the proposed model outperforms most of the other existing models in ZSL. Specifically, we outperform in AWA2,  CUB, and FLO while attaining comparable performance for other datasets. On the other hand, in case of GZSL, our structure aligning constraint helps the model to perform better in case of unseen test samples. This can be observed from the Table \ref{table:inductive-zsl-gzsl1} and \ref{table:inductive-zsl-gzsl2} where out model outperforms in all the datasets except AWA2. Specifically, we consistently perform better as compared to other methods in case of coarse-grained as well as fine-grained datasets. In the case of CUB and FLO, we gain by a large margin of 5.9 and 18, respectively, in the case of $H$. \textit{We note that the proposed model performance is compared with the performance of existing non-generative models only.} This is because most of the GZSL techniques using generative models make use of data augmentation technique. They generate unseen class data and then train supervised classifier over seen and generated unseen data. Furthermore, the superior performance exhibited by such models over the non-generative models can primarily be  attributed to an additional supervised classifier that is trained. While class labels are inferred using the trained supervised classifier, they differ from the standard GZSL protocol of predicting the class label using 1-nearest neighbor (1-NN) criterion.
\textbf{Transductive setting:} The results for transductive setting are given in Table \ref{table:transductive-gzsl} and \ref{table:transductive-zsl}.
Methods in QFSL (\cite{qfsl}) and SABR (\cite{iitropar}) although performs better than our model, they are not directly comparable as they use unseen class prototypes (unlabeled) during the training of the model. On the other hand, we stick to the standard transductive setting in (G)ZSL, which allows using only unlabeled unseen class visual features. We achieve the best performance using class-encoder instead of standard AE structure in the visual-latent-visual path of our model, similar to the inductive setting. As fewer training samples are available for CUB, class-encoder allows capturing better intra-class variance (by reconstructing each sample from every other sample of the same class) as compared to the standard AE.\\
For GZSL experiment using attributes, our model achieves the best performance on AWA2, CUB, and SUN in a similar experimental setting. We strongly outperform over the previous state of the art GFZSL (\cite{piyush-rai-expo}) by a considerable margin of around 38, 20 in AWA2 and CUB, respectively. Although GZSL poses a greater difficulty to model training due to model bias towards seen classes, using transductive setting helps in overcoming this bias issue. This can be observed from the performance on unseen class test samples given by $U$ in the Table \ref{table:transductive-gzsl} which is comparable with the performance over seen class test samples given by $S$. Furthermore, as compared to the inductive setting, an overall boost can be observed in $H$. From inductive to transductive H increases by 29.2 (AWA1), 33.3 (AWA2), 14.1 (CUB) and 19.6 (APY) thanks to unlabeled data and our self taught learning. Fig.\ref{fig:self-taught} shows t-SNE plots where the behavior of unseen class embeddings throughout training for AWA1 dataset can be observed. As training progresses, pseudo-labeled clusters in the initial steps get refined as well as become discriminative. The well clustered as well as discriminative embeddings are result of the self-taught learning of clusters using latent classifiers.
\begin{figure*}
\centering
   \includegraphics[scale=0.8, trim={0 0 0 0}]{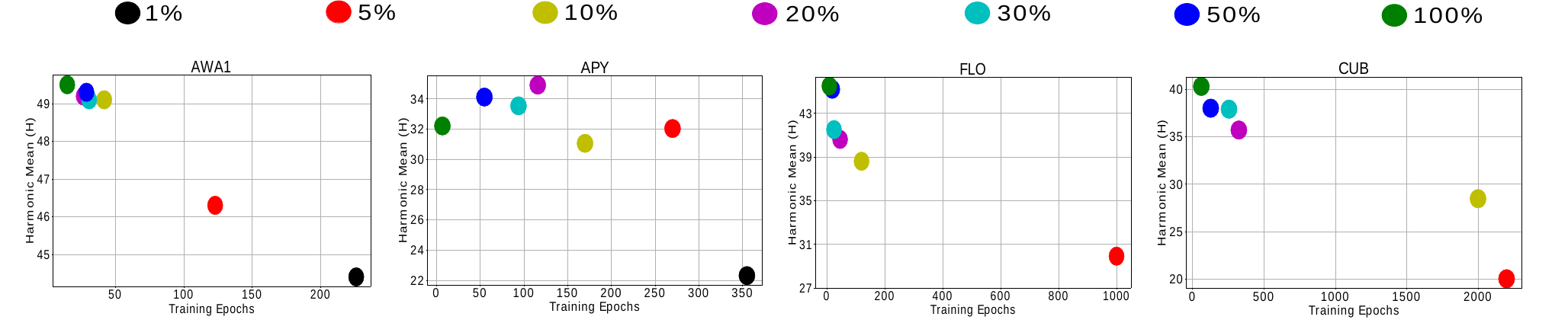}
\caption{Plots showing the effect on harmonic mean $(H)$ when the fraction of labeled seen data is used for training the GZSL model (in the inductive setting). Vertical axis represents Harmonic mean $H$. Horizontal axis represents training epochs. It can be observed that the proposed model performs well, even when using fewer samples per seen class for training. Each colored blob corresponds to $H$ given by the model trained on fraction (or $\%$) of seen data. Best viewed in color.}
\label{fig:lessdata}
\end{figure*}
\subsection{Extremely Less Labeled Data Regime}
 We recall that (G)ZSL comes in the less labeled data regime setting. It is interesting to evaluate how the proposed model performs when trained on extremely less labeled data in the inductive setting. To gain an insight into this, we further extend the less labeled data regime by using very few samples of seen classes for training the model. Fig.\ref{fig:lessdata} shows the performance of our model for subsets of training data. We experiment on two fine-grained datasets CUB and FLO and two coarse-grained datasets AWA1 and APY. We train our model using randomly selected $1\%$, $5\%$, $10\%$, $20\%$, $30\%$, and $50\%$ of labeled seen class data samples and record the harmonic mean $H$ in each case. For consistency and to avoid data imbalance, we use the same percentage of samples from each class. We note that for AWA1, the performance gap between $1\%$ and $100\%$ training data is only $5\%$. In the case of APY, our model behaves better when using less labeled data which shows the presence of outlier samples and suggest the need for outlier removal. In case of fine-grained FLO dataset, $H$ dips from 45.5 (100\% data) to 38.6 (10\% data), 40.6 (20\% data), 41.5 (30 \% data) , and 45.2 (50\% data). In case of fine-grained CUB dataset, $H$ dips from 40.3 (100\%) to 20.0 (10\% data), 28.45 (20\% data), 35.7 (30\% data), 37.9 (40\% data) and 38.0 (50\% data). 
\begin{table}
\caption{\label{tab1}Comparison of the performance in terms of the harmonic mean $H$ of our method against RN (\cite{learning-to-compare}) for different settings of extremely less labeled data regime. Numbers in the bracket represent the deviation from the average $H$ reported.}
\centering
\small
\setlength{\tabcolsep}{3.5pt}
\begin{tabular}{|c|c|c|c|c|}
\hline
 \textbf{Training Data} & \multicolumn{2}{|c|}{\textbf{AWA1}} & \multicolumn{2}{|c|}{\textbf{CUB}} \\
\cline{2-5}
\textbf{(\%)} & \multicolumn{1}{|c|}{Ours} & \multicolumn{1}{|c|}{RN} & \multicolumn{1}{|c|}{Ours} & \multicolumn{1}{|c|}{RN} 
\\
\hline 
1 & 44.4 (1.1) & 33.4 (4.2) & - & - \\
5 & 46.3 (0.8) & 33.8 (5.1) & 20.2 (0.7) & 12.5 (0.4) \\
10 & 49.1 (0.5) & 34.8 (5.5) & 28.4 (0.3) & 20.2 (1.1) \\
20 & 49.2 (0.6) & 36.4 (5.7) & 35.7 (0.6) & 28.7 (0.9) \\
30 & 49.1 (0.6) & 38.1 (6.2) & 37.9 (0.6) & 32.3 (0.8)\\
50 & 49.3 (0.4) &  42.7 (4.0)& 38.0 (0.2) & 37.8 (0.5)\\
\hline
\end{tabular}
\label{table:fractional-data}
\end{table}
We hypothesize that this behavior is due to our class-encoder based encoder-decoder framework which provides robust latent codes and structure aligning constraint used in training. The class-encoder aims to reconstruct a sample of the given class from a different sample of the same class. In such a case, even when a few training samples are available per class, it can generate multiple input-output pair for class-encoder training. This significantly alleviates the need for using a large number of training samples. On the other hand, as we align latent representations of the visual and semantic mean (Eq.(\ref{eqn:sa})), it eliminates the need for using a large number of training samples per class as long as classes are well clustered. While in the case of fine-grained datasets, training is inherently difficult, the effect of class-encoder based training on very few sample is not prominent.   

We note that we train the model on an average six samples (1\% data in AWA1), three samples (1\% data in APY), three samples (5\% data in CUB), and four samples (5\% data in FLO) per seen class. As the aforementioned experimental setting can also be treated as a few shot learning (\cite{one-shot}) in the context of ZSL, we compare our model with the few shot model proposed in (\cite{learning-to-compare}) on coarse-grained AWA1 dataset and fine-grained CUB dataset. We specifically choose the model in (\cite{learning-to-compare}) for the comparison as it is easy to evaluate and gives excellent performance in case of standard few-shot learning setting and is also used in ZSL. It can be observed from Table \ref{table:fractional-data} that our model outperforms model in (\cite{learning-to-compare}) in all the cases from $1\%$ to $50\%$ of seen data. This could be explained in the following way. The model in (\cite{learning-to-compare}) learns an embedding function by comparing samples from different classes. In such a case, any outlier samples from a randomly selected subset of training data can have an adverse effect on learning the embedding function. This can also be observed from the large deviation in the $H$ in the case of (\cite{learning-to-compare}). While (\cite{learning-to-compare}) uses meta-learning strategy based on episodic training we follow conventional regressor-driven ZSL setting. On the other hand, as our model makes use of encoder-decoder framework, its latent layer representations are more robust. Further, any outlier in the randomly selected samples would have little impact on learning of embedding functions as we try to align the means of visual and semantic samples. In such cases, outlier will have little impact on the mean as long as majority of the samples are well behaved. 
\section{Conclusions}
In this work, we proposed a shallow, but effective neural network-based model for (G)ZSL, which aligns class-neighborhoods in the latent embedding space while simultaneously making embedding space discriminative. The proposed model is evaluated in both inductive and transductive setting for ZSL and GZSL. While our proposed methodology to generate and use per sample semantic features using unseen class visual samples gives superior performance, it can in principle be used with any of the existing (G)ZSL models to reduce the model bias. Further, self-taught learning during the training regime helps in boosting the overall performance. We also show the superior performance of our model in case of extremely less labeled data regime where our model gives consistent recognition performance even when trained using a very few samples from seen classes.

{\small
\bibliographystyle{ieee_fullname}
\bibliography{egbib}
}

\end{document}